\documentclass[11pt]{article}
	
	\newcommand{\blind}{0}
        \newcommand{\vect}[1]{\boldsymbol{#1}}
	
    \makeatletter
    \renewcommand\section{\@startsection {section}{1}{\z@}%
                                       {-3.5ex \@plus -1ex \@minus -.2ex}%
                                       {2.3ex \@plus.2ex}%
                                       {\normalfont\fontfamily{phv}\fontsize{16}{19}\bfseries}}
    \renewcommand\subsection{\@startsection{subsection}{2}{\z@}%
                                         {-3.25ex\@plus -1ex \@minus -.2ex}%
                                         {1.5ex \@plus .2ex}%
                                         {\normalfont\fontfamily{phv}\fontsize{14}{17}\bfseries}}
    \renewcommand\subsubsection{\@startsection{subsubsection}{3}{\z@}%
                                        {-3.25ex\@plus -1ex \@minus -.2ex}%
                                         {1.5ex \@plus .2ex}%
                                         {\normalfont\normalsize\fontfamily{phv}\fontsize{14}{17}\selectfont}}
    \makeatother
	
	\usepackage{enumerate}
        \usepackage{natbib}
	\usepackage{url} 
        \usepackage[margin=1in]{geometry}

        \usepackage{graphicx}
        \usepackage{float}
        \usepackage{ctable}
        \usepackage{wrapfig}
        \usepackage{amssymb}
        \usepackage{multirow}
        \usepackage{amsfonts}
        \usepackage{amsmath}
        \usepackage{booktabs}
        \usepackage{makecell}
        \usepackage{array}
        \usepackage{amssymb}
        \usepackage{amsthm}
        \usepackage{booktabs}
        \usepackage{algorithm}
        \usepackage[noend]{algpseudocode}
        \usepackage{xcolor}
	
\begin{document}
		
		\def\spacingset#1{\renewcommand{\baselinestretch}%
			{#1}\small\normalsize} \spacingset{1}
		
		\if0\blind
		{
			\title{Transformer with Koopman-Enhanced Graph Convolutional Network for Spatiotemporal Dynamics Forecasting}
			\author{Zekai Wang $^a$ and Bing Yao $^b$\thanks{Corresponding author: byao3@utk.edu}  \\
			$^a$ Charles F. Dolan School of Business, Fairfield University, Fairfield, USA \\
            $^b$ Department of Industrial \& Systems Engineering, The University of Tennessee, Knoxville, USA }
			\date{}
			\maketitle
		} \fi

		\if1\blind
		{

            \title{\bf \emph{IISE Transactions} \LaTeX \ Template}
			\author{Author information is purposely removed for double-blind review}
			
\bigskip
			\bigskip
			\bigskip
			\begin{center}
				{\LARGE\bf \emph{IISE Transactions} \LaTeX \ Template}
			\end{center}
			\medskip
		} \fi
		\bigskip
		
	\begin{abstract}
Spatiotemporal dynamics forecasting is inherently challenging, particularly in systems defined over irregular geometric domains, due to the need to jointly capture complex spatial correlations and nonlinear temporal dynamics. To tackle these challenges, we propose TK-GCN, a two-stage framework that integrates geometry-aware spatial encoding with long-range temporal modeling. In the first stage, a Koopman-enhanced Graph Convolutional Network (K-GCN) is developed to embed the high-dimensional dynamics distributed on spatially irregular domains into a latent space where the evolution of system states is approximately linear. By leveraging Koopman operator theory, this stage enhances the temporal consistency during the latent learning. In the second stage, a Transformer module is employed to model the temporal progression within the Koopman-encoded latent space. Through the self-attention mechanism, the Transformer captures long-range temporal dependencies, enabling accurate forecasting over extended horizons. We evaluate TK-GCN in spatiotemporal cardiac dynamics forecasting and benchmark its performance against several state-of-the-art baselines. Experimental results and ablation studies show that TK-GCN consistently delivers superior predictive accuracy across a range of forecast horizons, demonstrating its capability to effectively model complex spatial structures and nonlinear temporal dynamics.
	\end{abstract}
			
	\noindent%

	\spacingset{1.5} 

\section{Introduction} \label{s:intro}
Spatiotemporal data characterize the evolution of dynamic processes across both spatial and temporal dimensions, offering critical insights into the behavior of complex systems \citep{wang2020deep, wikle2023statistical,yan2018real}. Advances in mobile sensing, remote imaging, and other monitoring technologies have substantially increased the volume, resolution, and accessibility of spatiotemporal datasets. These technological developments create unprecedented opportunities for data-driven modeling and forecasting of system dynamics. In biomedical domains, for instance, precise modeling of electrical wave propagation in cardiac tissue is vital for predicting arrhythmic events, enabling timely clinical interventions, and ultimately improving patient outcomes \citep{yang2023sensing}. Despite these opportunities, developing predictive models that effectively capture the intrinsic spatial and temporal structures inherent in spatiotemporal data remains fundamentally difficult with the following research challenges:

(1) \textbf{Complex Geometry and Spatial Structure}: Many real-world spatiotemporal systems are embedded in domains with irregular, non-Euclidean geometries that challenge conventional modeling approaches \citep{jin2024survey,zhao2021intrinsic}. For instance, the anatomical structure of the human heart involves intricately curved surfaces \citep{yao2016physics}, which defy representation by regular Cartesian grids. Accurately modeling such geometries is essential for predicting the propagation of electrical activation in cardiac tissues \citep{zhang2025geometry,yao2025simulation}. Similarly, urban transportation systems are naturally modeled as graphs, where intersections and road segments form nodes and edges with heterogeneous spatial arrangements, variable connectivity, and non-uniform distances \citep{yu2017spatio}. These structural complexities lead to spatial relationships that cannot be effectively captured by traditional Euclidean convolutions or grid-based methods. Robust forecasting models need to incorporate spatial representations that can flexibly encode both the geometric and topological characteristics of the domain to preserve locality, continuity, and relational structure inherent in the underlying system.

(2) \textbf{Nonlinear dynamics}: Spatiotemporal systems typically exhibit complex, nonlinear behaviors that emerge from underlying biological, environmental, or social processes, which simple linear models are fundamentally incapable of capturing \citep{liu2018statistical,yao2021spatiotemporal}. For example, in cardiac electrophysiology, the propagation of electrical impulses through excitable tissue can produce nonlinear phenomena such as reentrant spiral waves and spatiotemporal chaos, which are central to arrhythmogenesis \citep{xie2022physics}. Similarly, atmospheric systems are governed by nonlinear interactions among dynamic variables such as temperature, pressure, and wind velocity, resulting in phenomena like turbulence, sudden regime shifts, and long-range dependencies that complicate forecasting \citep{xu2021spatiotemporal}. Accurately modeling nonlinear dynamics demands expressive data-driven frameworks that can learn highly complex, non-additive mappings between past and future system states. 

(3) \textbf{Long-range temporal dependencies}: Spatiotemporal processes often involve long-range temporal dependencies, where system states at a given time are influenced by events that occurred far in the past. These dependencies are particularly critical in biomedical applications. For instance, in cardiac dynamics, a premature electrical stimulus (e.g., a premature ventricular contraction) may appear benign initially but can trigger complex arrhythmic events (e.g., tachycardia or fibrillation) several cycles later due to delayed interactions with tissue refractoriness and conduction pathways \citep{wit1990cellular,xie2024automated}. Effectively modeling such effects requires learning temporal patterns that span long time horizons. However, conventional sequence modeling architectures, e.g., recurrent neural networks (RNNs) and gated recurrent units (GRUs), often struggle to retain information over extended sequences due to vanishing gradients and cumulative prediction errors \citep{pascanu2013difficulty,wang2024muse,wang2021multi}. These limitations become more pronounced in real-world settings, where the complexity of temporal interactions is compounded by spatial heterogeneity and nonlinear dynamics. Overcoming these challenges demands advanced temporal modeling strategies that can maintain long-term memory and preserve signal fidelity across time scales.

In this paper,  we propose TK-GCN, a novel two-stage predictive framework for spatiotemporal dynamics forecasting that integrates a Transformer architecture with a Koopman-enhanced Graph Convolutional Network (K-GCN). In the first stage, K-GCN operates on graph-based representations of irregular spatial domains to model spatial dependencies in dynamic processes. Simultaneously, it leverages Koopman operator theory to embed the system dynamics into a latent space where they evolve in an approximately linear manner, facilitating more tractable temporal modeling. In the second stage, a Transformer is applied to the Koopman-encoded latent sequences, using self-attention to capture long-range temporal dependencies. By effectively integrating the spatial and temporal modeling, the proposed TK‐GCN model is capable of capturing complex nonlinear interactions, enabling more accurate information propagation across both dimensions. As a motivating case study, we evaluate our framework on cardiac spatiotemporal modeling to forecast the spread of electrical activation over a 3D heart geometry. Experimental results and ablation studies demonstrate that TK-GCN consistently outperforms state-of-the-art baselines in prediction accuracy and robustness. 

\section{Research Background}

 \subsection{Classical Statistical Approaches for Spatiotemporal Prediction}
  Spatiotemporal prediction \citep{wikle2023statistical,liu2022statistical} has been widely approached with statistical and geostatistical methods that extend time-series and spatial interpolation techniques. Spatiotemporal kriging, for example, is a geostatistical interpolation method that combines spatial and temporal covariance-based models to predict dynamic phenomena across space and time \citep{nag2023spatio}. Specifically, one typically assumes an underlying random field with a joint space-time covariance function. Given observations at various locations and times, this method produces an unbiased linear estimator for an unobserved location and time. Spatiotemporal kriging methods have been applied across various domains such as environmental monitoring, traffic engineering, and agriculture to interpolate and forecast processes that evolve over space and time \citep{graler2016spatio, yang2018kriging, snepvangers2003soil,wang2019modeling}. Although effective in many settings, spatiotemporal kriging struggles with large datasets due to its high computational cost and requires careful specification of the covariance structure, which may not capture nonstationary or anisotropic dynamics, making it less suitable for predictive modeling of complex or nonlinear systems.

Gaussian processes (GPs) \citep{rasmussen2003gaussian} offer a powerful Bayesian non-parametric framework for spatiotemporal modeling and prediction. Specifically, GPs represent spatial or spatiotemporal fields through covariance relationships between observations and enable automatic learning of kernel hyperparameters by optimizing the marginal likelihood or using Bayesian inference. In spatiotemporal settings, GPs require the specification of covariance kernels that encode dependencies across both spatial and temporal dimensions. The flexibility to design and combine kernels allows GPs to represent a broad range of correlation structures, including spatiotemporal composite patterns. This modeling capability has led to successful applications in various systems \citep{ba2012composite,singh2010modeling, jiang2022deep, senanayake2016predicting,xie2024hierarchical, tang2024hierarchical}.
Despite these strengths, GPs face several critical limitations in real-world spatiotemporal applications. First, their computational complexity scales cubically with the number of training points, which severely hampers scalability for large, high-resolution datasets. Although sparse and approximate GP variants exist \citep{hensman2015mcmc,datta2016hierarchical}, they often sacrifice accuracy or require intricate model design. Second, the expressiveness and generalization ability of a GP are fundamentally constrained by the choice of kernel. Designing effective kernels for systems with non-Euclidean or irregular geometries is particularly challenging \citep{calandra2016manifold}. Third, GP-based models often struggle with long-range temporal forecasting, especially in the presence of nonlinear dynamics, where the assumption of smooth, stationary covariance structures becomes invalid. 

\subsection{Deep Learning for Spatiotemporal Predictive Modeling}

Deep learning has been widely used as a powerful tool for spatiotemporal modeling, largely due to its capacity to capture complex nonlinear dependencies and scale effectively with large datasets. 
Early deep learning models for spatiotemporal data extend RNNs by integrating convolutional layers \citep{shi2015convolutional}. The key idea is to exploit convolution to capture local spatial correlations in each frame (or time step) while using the RNN or Long Short-Term Memory (LSTM) unit to capture temporal dependencies across frames. A seminal example is the ConvLSTM model proposed for precipitation nowcasting \citep{shi2015convolutional}, which allows the model to learn how spatial patterns such as rain fronts on a map move and evolve over time. Similar models have also been applied in areas such as urban and rural planning \citep{majidizadeh2024semantic}, transportation systems \citep{kwak2025hybrid}, and computer vision \citep{wang2020spatio}. 
However, ConvLSTM models are designed to Euclidean grids and face several challenges when applied to spatiotemporal forecasting on irregular domains. They assume fixed, regular neighborhoods, which breaks down on graphs or meshes, making it difficult to encode complex connectivity patterns. Moreover, ConvLSTMs rely on recurrent updates that can be computationally expensive on high-resolution spatial data and may suffer from vanishing gradients over long sequences, limiting their ability to produce reliable long-term forecasts.

Graph convolutional networks (GCNs) have emerged as a powerful tool for modeling spatial dependencies in data defined over irregular geometries and non-Euclidean domains \citep{bronstein2017geometric}. By extending the notion of convolution to graph-structured data, GCNs leverage the adjacency matrix or graph Laplacian to perform localized feature aggregation, capturing neighborhood-level spatial relationships. In spatiotemporal modeling, GCNs are often combined with temporal sequence models to jointly learn spatial and temporal dynamics. For instance, Li et al. \citep{li2017diffusion} introduced the diffusion convolutional RNN (DCRNN) for traffic forecasting, where traffic sensors form a graph over the road network. DCRNN employs diffusion-based graph convolutions to encode spatial dependencies and integrates them with a recurrent encoder-decoder architecture to model temporal evolution. Alternatively, Yu et al. \citep{yu2017spatio} proposed the spatiotemporal GCN (STGCN), a fully convolutional model that uses graph convolutions for spatial structure and 1D temporal convolutions to model temporal dependencies. By avoiding recurrent units, STGCN improves computational efficiency and training stability while achieving competitive performance.

Subsequent advances have extended spatiotemporal GCN modeling in various directions. Temporal attention mechanisms have been incorporated to dynamically weigh temporal inputs \citep{guo2019attention, li2022spatial}; hybrid models combining GCNs with LSTMs have been developed to capture sequential patterns \citep{gao2021graph, zhou2023grlstm}; and multi-graph architectures have been introduced to simultaneously model diverse spatial relationships such as topology, distance, and functional similarity \citep{geng2019spatiotemporal, shao2022long}. Despite their empirical success, current GCN-based spatiotemporal models face notable limitations. In particular, their reliance on localized graph convolutions hinders the ability to capture long-range spatial dependencies, especially in large or sparsely connected graphs. Moreover, temporal modules such as RNNs or CNNs often struggle to model complex nonlinear dynamics over extended horizons, leading to degraded accuracy in long-term forecasting.

More recently, the success of Transformer architectures \citep{vaswani2017attention, geneva2022transformers} in sequence modeling has started to influence spatiotemporal prediction models. Transformers rely on self-attention mechanisms to capture dependencies in sequences, advantageous for long sequences because they do not suffer from the step-by-step recurrence in RNNs that often leads to vanishing gradients and loss of long-term information.  In the context of pure time-series forecasting (without explicit spatial structure), Transformer-based models have achieved state-of-the-art results in long-term forecasting tasks \citep{wang2024muse,zhou2021informer, wen2022transformers}. Building on these advances, spatiotemporal Transformer models \citep{yan2021learning} have been proposed to handle both spatial and temporal dependencies. However, the majority of existing approaches are designed for data defined on regular Euclidean grids, which limits their ability to generalize to systems with irregular spatial domains or complex topologies. Recent developments, Graph Transformers, have attempted to address this limitation by introducing spatial inductive biases using graph-based attention mechanisms \citep{xu2020spatial, zheng2020gman}. Despite these efforts, current models often struggle to capture multiscale spatial correlations and long-range temporal dependencies simultaneously, especially in scenarios involving non-Euclidean 3D meshed geometries. Addressing these limitations remains a critical challenge for advancing Transformer-based models in real-world spatiotemporal systems.

\section{Research Methodology} 


We define the spatiotemporal prediction task over a spatial domain $\Omega$, which is discretized into $N$ distinct locations (or nodes), represented as the set $\mathcal{V} = \{1, 2, \dots, N\}$. Each location $i \in \mathcal{V}$ is associated with a state variable $x_i(t)$ that evolves over time $t$. For example, in cardiac electrodynamic modeling, $x_i(t)$ may represent the transmembrane voltage at the $i$-th node at time $t$. We define the state vector over the entire domain at time $t$ as $\vect{x}(t) = [x_1(t), x_2(t), \dots, x_N(t)]^\top \in \mathbb{R}^N$, which captures the system configuration (e.g., the voltage distribution over the heart) at that time. Assume we have discrete time steps, indexed by $t=1,2,\dots$.  Given the historical state trajectories up to time $T$, denoted by $\vect{x}(1), \vect{x}(2), \dots, \vect{x}(T)$, the objective is to forecast the future states over a prediction horizon of $\tau$ time steps, corresponding to time indices $T+1, T+2, \dots, T+\tau$. Formally, the goal is to learn a predictive mapping $\mathcal{F}$ such that, for each forecast step $h$ ($1 \leq h \leq \tau$), the future state $\vect{x}(T+h)$ is estimated as:
\begin{eqnarray}
\hat{\vect{x}}(T+h) = \mathcal{F}(\vect{x}(1),\vect{x}(2), \cdots, \vect{x}(T); h)
\end{eqnarray}
where $\hat{\vect{x}}(T+h)$ is the predicted state at time $T+h$. In other words, the function $\mathcal{F}(\cdot;h)$ maps the sequence of past system states to a prediction of the future state $h$ steps ahead. 

\begin{figure}[!ht]
	\begin{center}
		\includegraphics[width=5.5in]{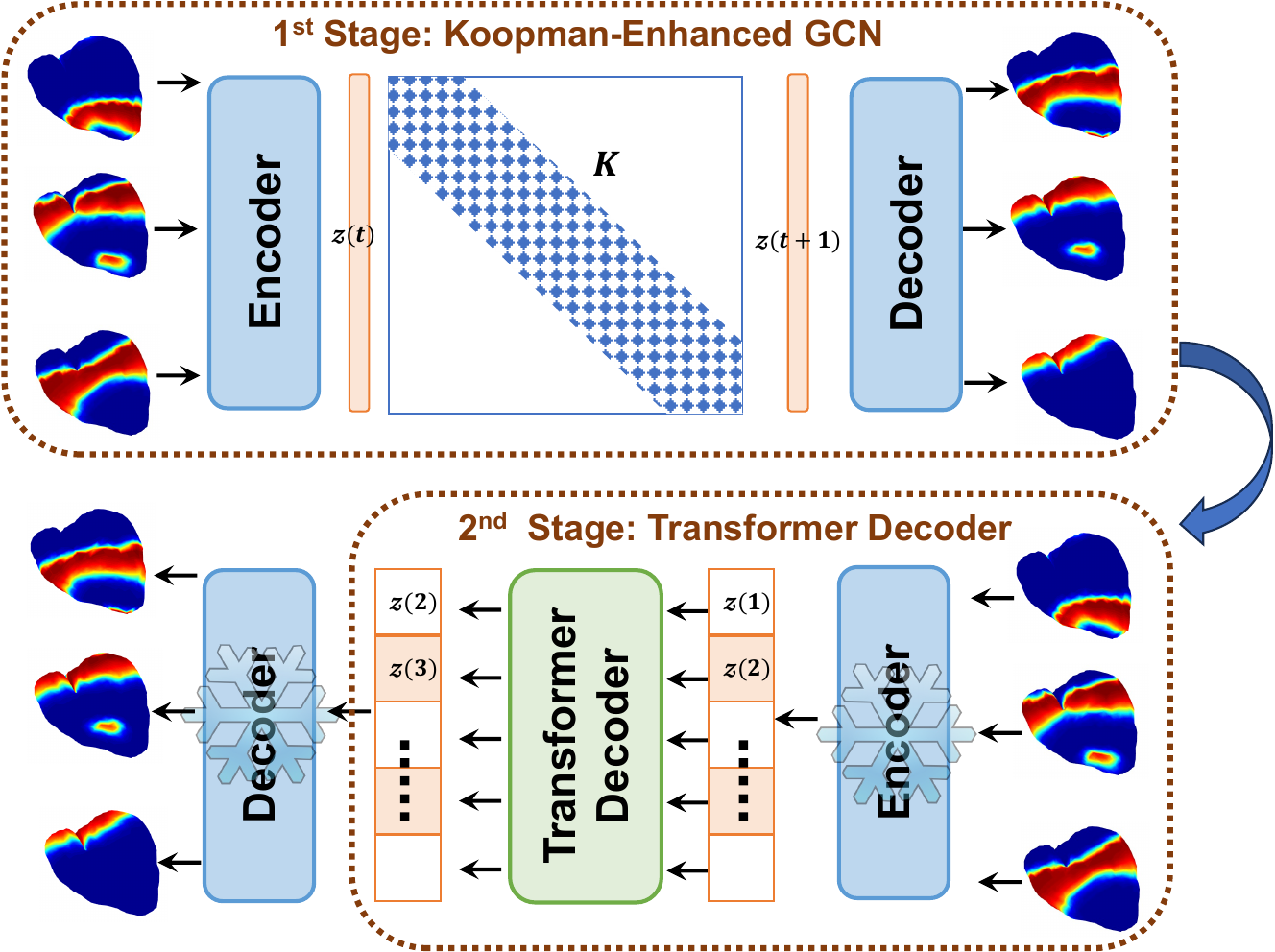}
		\caption{Flowchart of the proposed two-stage TK-GCN framework: In the first stage, we propose a Koopman-Enhanced Graph Convolutional Network (K-GCN) to establish the Encoder-Decoder framework and learn a latent representation of the spatiotemporal dynamics. In the second stage, the pretrained Encoder is frozen and repurposed as a fixed embedding module. A Transformer-based decoder is then employed to model the temporal evolution within the learned latent space for future state forecasting. }
		\label{Fig:TwoStage}
	\end{center}    
 
\end{figure}

However, learning $\mathcal{F}$ for spatiotemporal systems is generally intractable due to the high dimensionality and nonlinear interactions of the underlying dynamics. To cope with the challenges, we propose a two-stage modeling approach. As shown in Fig. \ref{Fig:TwoStage}, we first develop a Koopman-Enhanced Graph Convolutional Network (K-GCN) to learn a low-dimensional latent representation in which the system dynamics are more amenable to modeling. Then, we propose to leverage the powerful learning capability of Transformer to capture the temporal evolution within this latent space and forecast future system states. The subsequent sections provide a detailed exposition of the proposed architecture and its individual components.

\subsection{Koopman-Enhanced Graph Convolutional Network (K-GCN)}

\begin{figure}[!ht]
	\begin{center}
		\includegraphics[width=6.5in]{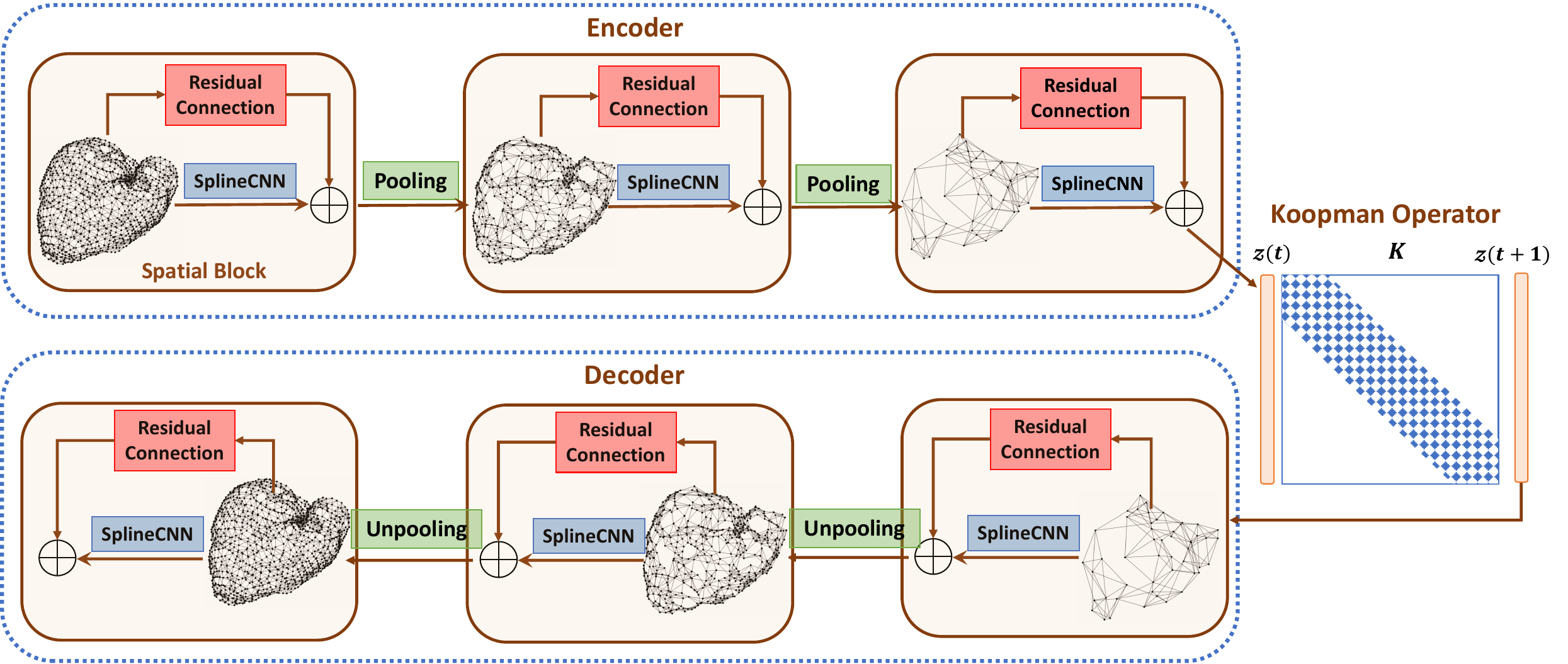}
		\caption{The first-stage K-GCN model to learn a low-dimensional latent representation.}
		\label{Fig:1st stage}
	\end{center}    
\end{figure}

As shown in Fig. \ref{Fig:1st stage}, the Stage-1 K-GCN model integrates spatial and temporal information by combining a Graph Convolutional Network (GCN), which captures the underlying geometric structure of the spatial domain, with a learnable Koopman operator that embeds the temporal evolution nature of the system state into the learning process. This design enables the extraction of an informative low-dimensional embedding of the spatiotemporal dynamics. The learned embeddings are then used as inputs to a Transformer-based sequence model, which captures long-range latent temporal dependencies (with the Encoder kept fixed) and forecasts future system states over time.

\subsubsection{Graph Convolutional Networks (GCNs)}

GCNs extend classical convolutional operations to irregular, graph-structured data by leveraging the inherent connectivity among nodes. We propose an advanced graph learning framework to account for the complex geometry in dynamic modeling by leveraging cutting-edge GCNs. Specifically, the 3D geometry is modeled as a single undirected graph $\mathcal{G} = ({\cal V}, {\cal E}, \mathbf{W}, \mathbf{X})$. Here, ${\cal V}$ is the set of $N$ nodes, and ${\cal E}\subseteq {\cal V}\times {\cal V}$ describes their connectivity with ${\cal E}_{ij}=1$ if nodes $i$ and $j$ are connected and 0, otherwise. The tensor $\mathbf{W} \in [0,1]^{N \times N \times 3}$ represents the edge attributes with the element $\vect{w}(i,j)\in \mathbb{R}^3$ defined as normalized coordinate differences between vertices $i$ and $j$ if ${\cal E}_{ij}=1$ and $\vect{0}$, otherwise. Meanwhile, $\mathbf{X}=\{\vect{x}(t)\}_{t=1}^T \in \mathbb{R}^{N \times M \times T}$ holds the time-series data for all vertices, where $M$ denotes the number of features per node, and $T$ is the number of temporal snapshots. This graph-based representation allows us to leverage graph convolution operations to learn rich spatial feature embeddings on the mesh. Specifically, this graph learning framework includes three key components.

\textbf{Spatial Graph Convolution}: The graph convolutional operation is defined to improve the network representation and training by leveraging the inherent spatial correlation of nearby nodes. Note that the spatial correlations of input node features are locally captured by the edge attribute $\mathbf{W}$. As such, we propose to leverage $\mathbf{W}$ and design the graph convolutional kernel to determine how the input features are aggregated for preserving key geometric information. Specifically, the convolution operates on elements $\vect{w}$ in $\mathbf{W}$ as 
     \begin{eqnarray}
         g_{\Theta}(\vect{w})=\sum_{\vect{p}\in \mathcal{P}}\theta_{\vect{p}}\cdot B_{\vect{p}}(\vect{w})
     \end{eqnarray}
   where $\Theta = \{\theta_{\vect{p}}\}_{\vect{p}\in \mathcal{P}}$ is the trainable parameters in the convolution operations, and $B_{\vect{p}}$ is the product of the basis function. In our framework, we use continuous B-spline kernels for convolution, following the SplineCNN approach \citep{fey2018splinecnn}. In B-spline learning, we let $((N_{1,i})_{1\leq i\leq k_1}, \allowbreak (N_{2,i})_{1\leq i\leq k_2}, \allowbreak (N_{3,i})_{1\leq i\leq k_3})$ denote the basis obtained from equidistant knot vectors with $\vect{k}=(k_1,\allowbreak k_2,\allowbreak k_3)$ specifying the kernel size for each coordinate dimension; $\mathcal{P}$ denotes the set of Cartesian product of the basis: $\mathcal{P} = \{(N_{1,i})_{1\leq i\leq k_1},(N_{2,i})_{1\leq i\leq k_2},(N_{3,i})_{1\leq i\leq k_3}\}$. Hence, we have 
     \begin{eqnarray}
         B_{\vect{p}} =\Pi_{i=1}^3N_{i,p_i}(w_i), ~~~~ \vect{w}=(w_1, w_2, w_3)\in \vect{W}, ~~~~ \vect{p}=(p_1,p_2,p_3)\in \mathcal{P}
     \end{eqnarray}
   
   Given the kernel function $g_{\Theta}(\cdot)$ and node features $\mathbf{X}$, the graph convolution operation for node $i$ is defined as
     \begin{eqnarray}
         \left(\mathbf{X}*g\right)_i = \frac{1}{|\mathcal{N}(i)|}\sum_{j\in \mathcal{N}(i)}\vect{x}_jg_{\Theta}(\vect{w}(i,j))
     \end{eqnarray}
    where $\mathcal{N}(i)$ is the local neighborhood of node $i$ on the geometry, and $\vect{x}_j$ is the feature vector of node $j$. This continuous kernel formulation (SplineCNN) enables the convolution to smoothly adapt to the irregular geometry of the mesh by interpolating features as a function of the edge’s pseudo-coordinates. As a result, the GCN can effectively learn translation- and rotation-invariant spatial filters on the mesh domain, analogous to classical CNN filters on regular grids.

\textbf{Residual Connection}: As shown in Fig. \ref{Fig:1st stage}, we adopt a residual connection to stabilize training and improve feature propagation in our K-GCN spatial block. Specifically, the input mesh features in each block are first passed through a SplineCNN layer to extract localized geometric information. In parallel, the original input is propagated through a residual connection and then combined with the SplineCNN output via element-wise addition. This additive skip connection allows the network to directly propagate the input features alongside the transformed ones, facilitating effective gradient flow and mitigating gradient degradation issues \citep{he2016deep}. The residual structure is expressed mathematically as:
\begin{eqnarray}
\mathbf{X}^{(l+1)} = \mathbf{X}^{(l)} + \text{SplineCNN}(\mathbf{X}^{(l)})
\label{residual connection}
\end{eqnarray}
where $\mathbf{X}^{(l)}$ denote the input node features at layer $l$. This residual structure enables each layer to focus on learning incremental updates to the input features, rather than computing entirely new feature representations from scratch, which enhances both model stability and generalization when stacking multiple graph convolutional layers.

\textbf{Graph Pooling and Unpooling}: To capture multi-scale spatial features and reduce computational complexity, we construct a hierarchical graph representation via graph coarsening to facilitate graph pooling. In our approach, the original graph $\mathcal{G}$ representing the 3D meshed geometry is first partitioned into disjoint clusters $\mathcal{C}_1,\dots,\mathcal{C}_{N_c}$ via the Graclus clustering algorithm \citep{dhillon2007weighted}, such that $\mathcal{G}=\cup_{i=1}^{N_c}\mathcal{C}_i$ and $\mathcal{V}_i$ denoting the vertices in cluster $\mathcal{C}_i$. Then, each cluster $\mathcal{C}_i$ is contracted into a single super-node, resulting a coarsened graph $\mathcal{G}_c = (\mathcal{V}_c,\mathcal{E}_c, \mathbf{W}_c, \mathbf{X}_c)$. The vertice coordinates of the coarsened graph $\vect{V}_c=[\vect{v}_{c,1};\dots;\vect{v}_{c,N_c}]\in \mathbb{R}^{N_c\times 3}$ are derived as
     \begin{eqnarray}
         \vect{V}_c = \Delta_cP\vect{V}
     \end{eqnarray}
    where $\vect{V}\in\mathbb{R}^{N\times 3}$ are the original vertice coordinates, $P \in \mathbb{R}^{N_c \times N}$ is a binary cluster assignment matrix defined as
\begin{eqnarray}
P_{ij} = 
\begin{cases}
1, & \text{if the node } j ~\text{in the original graph} \text{ belongs to cluster } \mathcal{C}_i, \\[1mm]
0, & \text{otherwise.}
\end{cases}
\end{eqnarray}
   and $\Delta_c\in \mathbb{R}^{N_c\times N_c}$ is a diagonal matrix with elements defined as $\Delta_{c,ii} = \frac{1}{\sum_{j=1}^N P_{ij}}$. This design ensures that the coarsened node position $\vect{v}_{c,i}$ is the centroid of the original nodes in cluster $\mathcal{C}_i$. The edge $\mathcal{E}_c$ is updated by deriving the k-nearest graph given the coarsened node coordinates $\vect{V}_c$, and the edge attributes $\vect{W}_c$ will be updated by calculating the normalized vector displacement between the nodes, $\mathcal{V}_c$, in the coarsened graph. Furthermore, the pooled feature matrix $\vect{X}_{c} \in \mathbb{R}^{N_c \times M\times T}$ of the coarsened graph is obtained by
\begin{eqnarray}
\vect{X}_{c} = \Delta_cP\,\vect{X},
\end{eqnarray}
 which aggregates features within each cluster, thereby producing a compact representation of the spatial data. This hierarchical pooling can be recursively applied, resulting in a progressive reduction of graph resolution and complexity, while preserving essential structural information.

To recover fine-scale representations from the coarsened graph, we implement an unpooling operation that maps features from coarse nodes back to their corresponding fine-scale nodes. 
Specifically, the unpooled features \( \mathbf{X}_{\text{unpool}} \in \mathbb{R}^{N \times M\times T} \) are derived from \( \mathbf{X}_{c} \in \mathbb{R}^{N_c \times M\times T} \) according to:
\[
\mathbf{X}_{\text{unpool}} = P^\top \mathbf{X}_c,
\]
   where \( P^\top \in \mathbb{R}^{N \times N_c} \) redistributes each coarse feature across its corresponding cluster of fine-scale nodes. This pooling–unpooling scheme ensures end-to-end differentiability and preserves multi-scale spatial fidelity in the learned representations.

\subsubsection{Koopman Operator for Spatiotemporal Dynamics}
    To effectively incorporate the dynamic evolution nature of the complex spatiotemporal systems, we propose to integrate the Koopman operator framework into the GCN for effective latent space learning. Koopman operator theory provides a linear perspective on nonlinear dynamic systems by operating on observables of the state $\vect{x}(t)$ rather than the state itself \citep{yeung2019learning,biehler2024detonate}. Consider a discrete-time nonlinear dynamic system where the state evolves according to $\vect{x}(t+1) = f(\vect{x}(t))$ with $f(\cdot)$ being a potentially nonlinear evolving function. The Koopman operator $\mathcal{K}$ is a linear operator acting on observable function $\phi(\vect{x})$ and satisfies:
       \begin{eqnarray}
         \mathcal{K} \phi(\vect{x}(t)) = \phi \circ f(\vect{x}(t)) \overset{\Delta}{=} \phi (f(\vect{x}(t)))=\phi(\vect{x}(t+1))
         \label{Eq: K}
      \end{eqnarray}
      
    Eq. (\ref{Eq: K}) indicates that although the underlying dynamics $f(\cdot)$ may be nonlinear, Koopman operator advances the observables linearly in time. Note that this operator is typically infinite-dimensional.   
    To make this approach practical, we propose to leverage the GCN latent encoding to find a suitable low-dimensional set of observables, and then the dynamics in the embedded space can be approximated by a finite matrix that behaves like the Koopman operator. Specifically, we introduce a trainable matrix $K \in \mathbb{R}^{d_z \times d_z}$ to serve as an approximate Koopman operator in the latent space, where $d_z$ is the dimension of the latent vector $\vect{z}(t)$, and  $\vect{z}(t)=\psi_e(\vect{x}(t))$ with $\psi_e(\cdot)$ denoting a series of encoding operations composed of graph convolutions and pooling. The goal is to enforce that the latent states evolve (approximately) linearly under $K$, i.e.,
    \begin{eqnarray}
        K\vect{z}(t) = K\circ \psi_e(\vect{x}(t)) \approx \psi_e(\vect{x}(t+1)) =\vect{z}(t+1)
    \end{eqnarray}
    By learning an encoding that respects this linear transition as much as possible, we embed the nonlinear dynamics of the original system into a latent space where the dynamics are more tractable to model. The next-step latent state $\vect{z}(t+1)$ will then go through a series of decoding operations $\psi_d(\cdot)$ composed of graph unpooling and convolutions back to the original state space to predict the dynamics at the next time step, i.e., $\hat{\vect{x}}(t+1)=\psi_d(\vect{z}(t+1))$.

The GCN-based encoder-decoder (i.e., $\psi_e$ and $\psi_d$) and the Koopman operator $K$ are trained together in an autoencoder fashion. 
 The overall loss function $L_{\mathrm{total}}$ is composed of three parts: (i) a reconstruction loss, $L_{\mathrm{recon}}$, to ensure the autoencoder accurately reconstructs the input, (ii) a Koopman dynamics loss, $L_{\mathrm{dyn}}$, to enforce latent linear prediction, and (iii) a regularization on the Koopman operator, $L_{\mathrm{decay}}$, for stability. In particular, given the dataset $ \big\{\mathbf{x}(t)\big\}_{t=1}^{T}$, we define each loss component as:
\begin{eqnarray}
L_{\mathrm{recon}} &=& \frac{1}{T}\sum_{t=1}^{T} \Bigl\|\, \vect{x}(t) \;-\; \psi_d\big(\psi_e\big(\vect{x}(t)\big)\big)\Bigr\|^2 \\
L_{\mathrm{dyn}}   &=& \frac{1}{T}\sum_{t=1}^{T-\Delta T}\sum_{\Delta t=1}^{\Delta T} \Bigl\|\,\vect{x}(t+\Delta t) \;-\;  \psi_d\big(K^{\Delta t}\cdot \psi_e\big(\vect{x}(t)\big)\Bigr\|^2 \\
L_{\mathrm{decay}}   &=& \bigl\|\,K\bigr\|_F^2
\end{eqnarray}
where $K^{\Delta t}\cdot \psi_e\big(\vect{x}(t)\big)=\psi_e \big({\vect{x}(t+\Delta t)}\big)$, which advances the latent vector $\Delta t$ steps ahead. Finally, the total loss given weight parameters $\lambda_1$ and $\lambda_2$ is defined as:
\begin{eqnarray}
L_{\mathrm{total}} &=& L_{\mathrm{recon}} + \lambda_1 L_{\mathrm{dyn}} + \lambda_2 L_{\mathrm{decay}}
\label{total loss}
\end{eqnarray}

\subsection{Transformer Architectures for Latent Embeddings Modeling.}
Transformers are well-suited for long-range temporal modeling: the self-attention mechanism allows the model to capture dependencies between distant time steps, overcoming the vanishing gradients that plague traditional recurrent networks. Moreover, the self-attention enables fully parallel sequence processing during training (unlike the step-by-step updates in RNNs), greatly improving efficiency. Due to these properties, in the second stage of our framework, we employ a Transformer decoder to generate future latent embeddings, given the sequence of encoded latent variables from stage-1 K-GCN.

\subsubsection{Transformer Decoder Architecture}
Our decoder follows the standard Transformer architecture \citep{vaswani2017attention, wolf2020transformers}. The input consists of multiple sequences generated by applying a sliding window of fixed length $L$ to the original dataset, denoted as: $\{Z_1, Z_2, Z_3, \dots\}$, where $Z_j = [\vect{z}(t_{j-L+1}), \vect{z}(t_{j-L+2}) \dots, \vect{z}(t_j)]^\top \in \mathbb{R}^{L \times d_z}$ consists of latent vectors learned from the first-stage K-GCN encoder. Since the Transformer architecture is neither recurrent nor convolutional, it does not inherently encode the sequential order of the input embeddings. To capture the relative positions of latent vectors in the sequence, we incorporate positional encodings using the standard sine and cosine functions:
\begin{eqnarray}
PE_{(t_j,2i)}   = \sin\Bigl(t_j/10000^{\frac{2i}{d_{z}}}\Bigr), \quad \quad PE_{(t_j,2i+1)} = \cos\Bigl(t_j/10000^{\frac{2i}{d_{z}}}\Bigr),
\label{PE}
\end{eqnarray}
for the $2i$ and $2i +1$ elements in the latent variables $\vect{z}(t_j)$. The positional embedding features are then combined with the original latent variables to form the input for the Transformer decoder: $Y_j = PE_j + Z_j$, where $PE_j$ is a matrix with elements defined in Eq. (\ref{PE}). 

\begin{figure}[!ht]
	\begin{center}
		\includegraphics[width=6.5in]{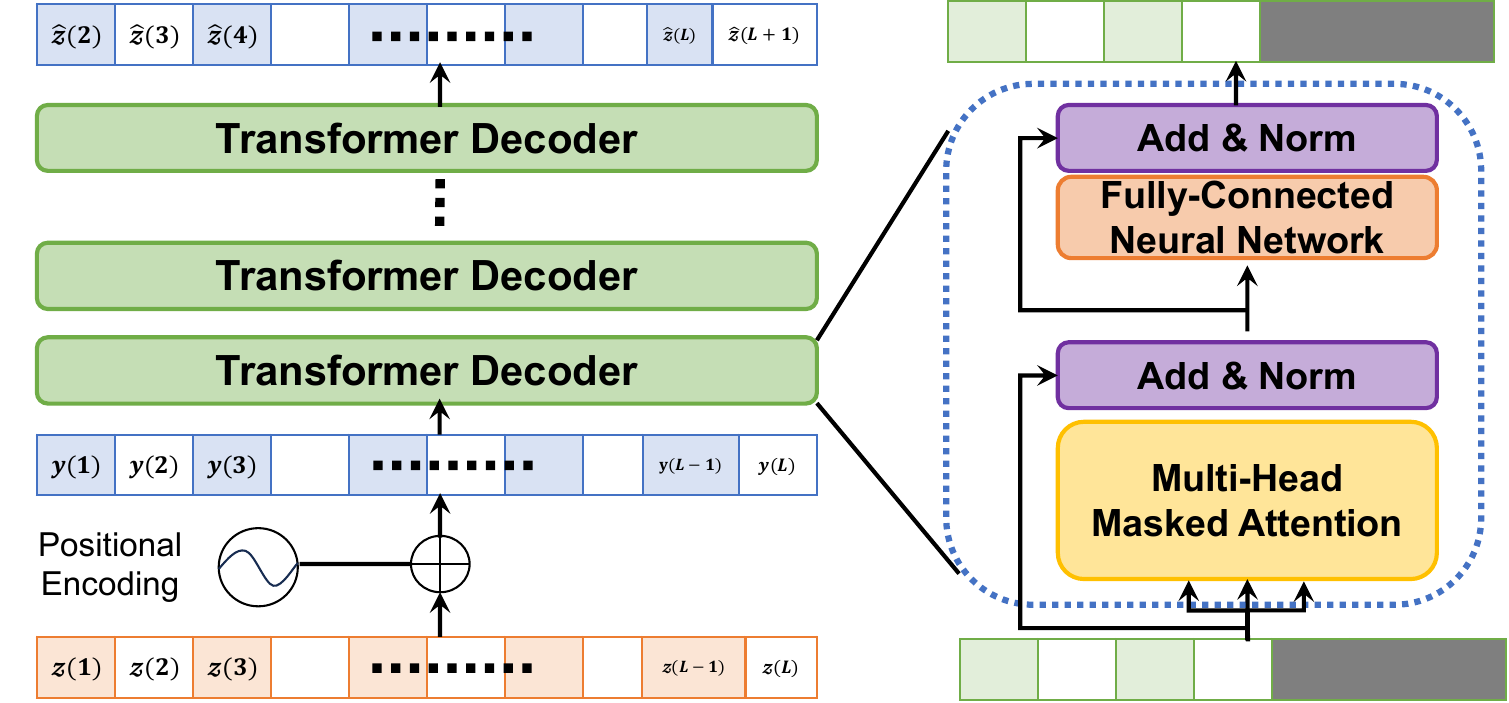}
		\caption{The Transformer decoder architecture for spatiotemporal dynamics forecasting.}
		\label{Transformer}
	\end{center}    
 
\end{figure}

As shown in Fig. \ref{Transformer}, the sequence $Y_j$
  is subsequently processed by a stack of Transformer decoder layers, each designed to model complex temporal dependencies in the output space while maintaining the autoregressive nature of the prediction task. Specifically, each decoder layer comprises two sub-components: (i) a masked multi-head self-attention mechanism that ensures information at each position can only attend to previous positions in the sequence, thereby preserving causality and preventing information leakage from future time steps; and (ii) a position-wise fully connected feed-forward network, typically consisting of linear transformations with a non-linear activation (e.g., ReLU), which enhances the model capacity for non-linear feature transformation. Residual connections and layer normalization are applied after each sub-layer to facilitate gradient flow during training and stabilize the learning dynamics.

 The multi-head attention mechanism deploys multiple attention heads in parallel, allowing the model to attend to different patterns or time-scales simultaneously. For each attention head $h \in\{1, 2, \cdots, H\}$, the corresponding query $Q^{(h)}$, key $K^{(h)}$, and value $V^{(h)}$ matrices are computed via learned projections applied to the input sequence $Y_j$:
\begin{eqnarray}
    Q^{(h)} = \mathcal{F}_Q^{(h)}(Y_j) \in \mathbb{R}^{L \times \frac{d_z}{H}}, \quad K^{(h)} = \mathcal{F}_K^{(h)}(Y_j) \in \mathbb{R}^{L \times \frac{d_z}{H}}, \quad V^{(h)} = \mathcal{F}_V^{(h)}(Y_j) \in \mathbb{R}^{L \times \frac{d_z}{H}},
    \label{QKV}
\end{eqnarray}
where $\mathcal{F}_Q^{(h)}(\cdot), \mathcal{F}_K^{(h)}(\cdot), \mathcal{F}_V^{(h)}(\cdot)$ represent the trainable network operations to calculate the query, key, and value matrices, respectively. The scaled masked attention for head $h$ is then given by:
\begin{eqnarray}
\mathrm{Attn}^{(h)}(Q^{(h)},K^{(h)},V^{(h)}) = \mathrm{softmax}\!\left(\frac{Q^{(h)} (K^{(h)})^\top}{\sqrt{d_k}} \odot C + (1 - C) \cdot B\right) V^{(h)}. 
    \label{Attn}
\end{eqnarray}
 where  $C\in \mathbb{R}^{L\times L}$ is a causal mask used to enforce autoregressive constraints. Specifically, $C_{ij}=1$ for $j< i$ and $C_{ij} = 0$ for $j\geq i$, and $B$ represents a large negative constant (e.g., $-10^{9}$), ensuring that the attention weights for future time steps are suppressed before applying the softmax. As a result, this mask ensures that each $\vect{z}(t_j)$ only attends to latent vectors at time steps before $t_j$, i.e., $\{\vect{z}(t_{j-L+1}), \cdots, \vect{z}(t_{j-1})\}$, thereby preserving the autoregressive property and preventing information leakage from future observations.

The attention outputs of all $H$ heads are first concatenated along the feature dimension and then projected through a learnable linear transformation to form the multi-head attention output:
\begin{eqnarray}
\mathrm{MultiHead}(Y_j) = \left[ \mathrm{Attn}^{(1)}; \, \mathrm{Attn}^{(2)}; \, \dots; \, \mathrm{Attn}^{(H)} \right] W_O
    \label{MultiHead}
\end{eqnarray}
with a learnable weight matrix $W_O \in \mathbb{R}^{d_z \times d_z}$. Then, the $\mathrm{MultiHead}(Y_j)$ is added to its original input to achieve the residual connection, followed by the layer normalization:
\begin{eqnarray}
    \tilde{Y_j} = \mathrm{LayerNorm}\Bigl( Y_j + \mathrm{MultiHead}(Y_j) \Bigr),
\label{residual_layer}
\end{eqnarray}
which is used to stabilize the training of a deep network. The normalized output $\tilde{Y_j}$ is then processed by a fully connected feed-forward network (FFN), resulting in $\mathrm{FFN}(\tilde{Y_j})$, and a second residual connection is applied followed by layer normalization to generate the final output of one transformer decoder block: 
\begin{eqnarray}
    \hat{Z}_{j+1} = \mathrm{LayerNorm}\Bigl( \tilde{Y_j} + \mathrm{FFN}(\tilde{Y_j}) \Bigr).
\label{second_residual_layer}
\end{eqnarray}
where the final output $\hat{Z}_{j+1}=[\hat{\vect{z}}(t_{j-L+2}), \hat{\vect{z}}(t_{j-L+3}) \dots, \hat{\vect{z}}(t_{j+1})]^\top \in \mathbb{R}^{L \times {d_z}}$ is a predicted sequence of latent vectors with the same shape as the input. The transformer block will be trained by minimizing the following forecasting loss function:
  \begin{eqnarray}
      L_{\text{fore}}=\sum_{j=L}^{T}\|\hat{Z}_{j+1}-Z_{j+1}\|^2=\sum_{j=L}^{T}\sum_{k=2}^L\|\hat{\vect{z}}({t_{j-k+2}})-{\vect{z}}({t_{j-k+2}})\|^2.
  \end{eqnarray}

  After training the Transformer-based forecasting model in the latent space, an unseen test sequence $Z_{t_0} = \left[\vect{z}(t_0 - L + 1), \vect{z}(t_0 - L + 2), \dots, \vect{z}(t_0)\right]^\top \in \mathbb{R}^{L \times d_z}
$
is provided as input to the trained model. The model then performs iterative multi-step forecasting to generate the future latent trajectory over a horizon of $\tau$ steps:
\begin{equation}
\hat{Z}_{t_0 + \tau} = \left[\hat{\vect{z}}(t_0 + 1), \hat{\vect{z}}(t_0 + 2), \dots, \hat{\vect{z}}(t_0 + \tau)\right]^\top \in \mathbb{R}^{\tau \times d_z},
\end{equation}
where each $\hat{\vect{z}}(t_0 + i)$ is recursively predicted by feeding the previously predicted latent vectors back into the model, preserving the autoregressive structure. Finally, the predicted latent sequence $\hat{Z}_{t_0+\tau}$ is passed through the GCN-decoder, which maps the latent representation back to the original input state space, yielding the final $\tau$-step-ahead forecast of the system dynamics:
\begin{equation}
\hat{X}_{t_0+\tau} = \left[\hat{\vect{x}}(t_0 + 1), \hat{\vect{x}}(t_0 + 2), \dots, \hat{\vect{x}}(t_0 + \tau)\right]^\top=\psi_d(\hat{Z}_{t_0+\tau}) \in \mathbb{R}^{\tau \times N}.
\end{equation}

\section{Experimental Design and Results} 

\subsection {Model Architecture}

\begin{figure}[!ht]
	\begin{center}
		\includegraphics[width=6in]{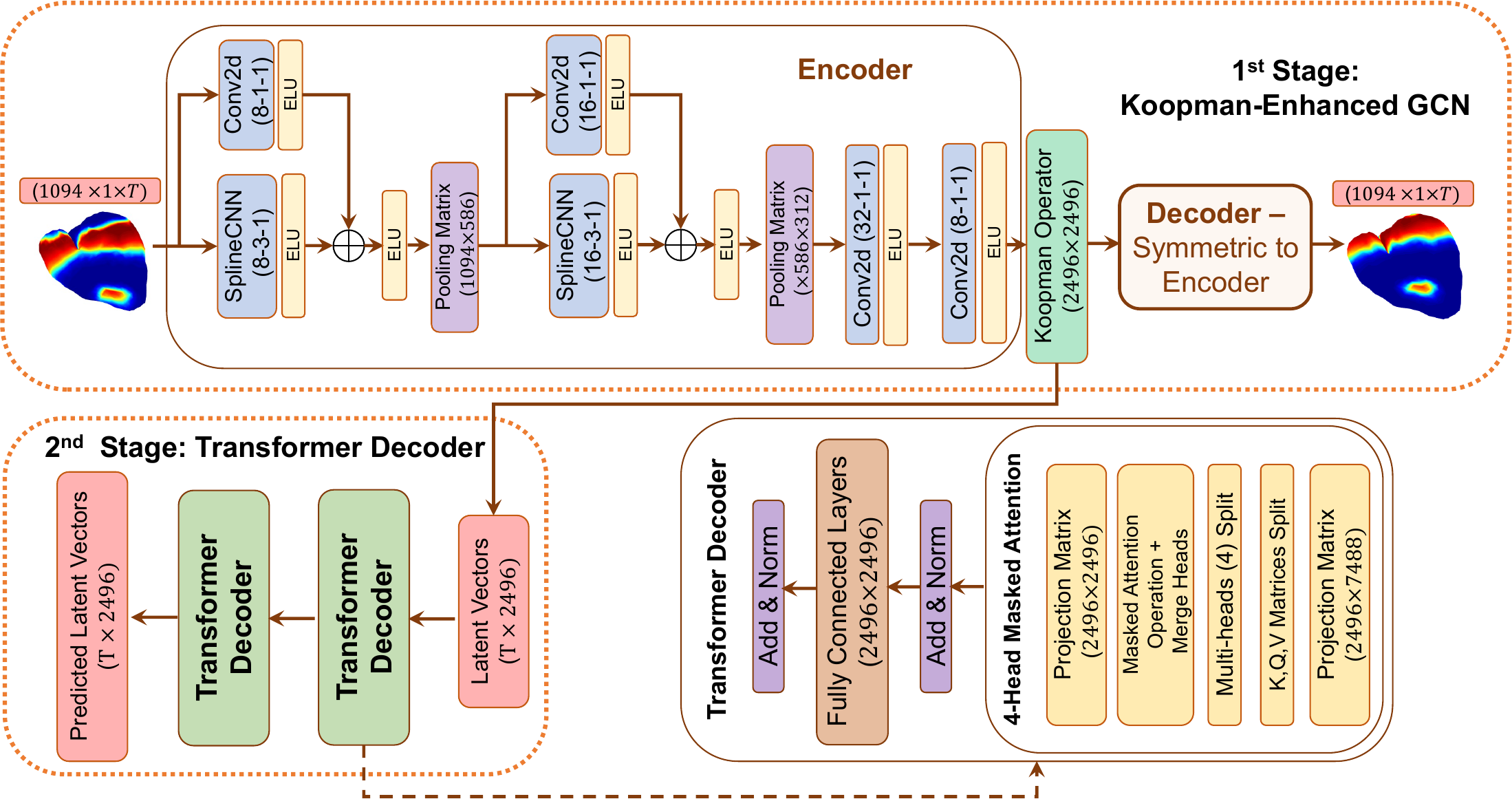}
		\caption{The architecture detail of the proposed TK-GCN framework}
		\label{Fig:arch details}
	\end{center}    
\end{figure}

Fig. \ref{Fig:arch details} illustrates the architecture of our TK-GCN framework. The stage-1 K-GCN follows an autoencoder structure with an encoder, a bottleneck latent layer, and a decoder. The encoder consists of two hierarchical spatial blocks followed by two pointwise convolutional layers. Each spatial block employs a SplineCNN layer with B-spline kernels and residual connections to preserve spatial features. The SplineCNN configuration, denoted as (8-3-1), indicates that the layer contains 8 filters, and employs a degree-1 B-spline basis with a kernel size of $k_1 = k_2 = k_3 = 3$. Similarly, the notation (8-1-1) for Conv2d layers refers to 8 filters with a kernel size of 1 and a stride of 1. Both SplineCNN and Conv2d are followed by an ELU activation \citep{clevert2015fast}. The input graph contains 1-D features per node, with a total of 1094 nodes. Between the Encoder and Decoder, we introduce a learnable Koopman operator $K$ to incorporate the dynamic evolution nature of the system into the latent learning. After applying the Koopman operator, the predicted latent vector is then propagated through the Decoder to reconstruct the system state at the next time step in the original space. The Decoder architecture is symmetric to the Encoder and is designed to invert the spatial coarsening and feature compression applied during encoding.

To model the temporal evolution within the learned latent space in Stage 2, we employ a Transformer decoder adapted from the GPT-2 model \citep{radford2018improving, radford2019language}. The architecture consists of two identical Transformer decoder layers, each equipped with a 4-head masked self-attention mechanism to capture temporal dependencies within the latent space. The input to the Transformer is a sequence of latent vectors over a sliding window $L$ (with $L = 128$ in our experiments), which is then projected into a higher-dimensional space using a learnable linear projection. The resulting representation is further structured to produce the concatenated query, key, and value matrices required for the multi-head attention mechanism. Each attention head receives its respective slice of the query and key matrices, enabling the model to attend to distinct subspaces of the temporal dependencies. The outputs from all attention heads are concatenated and passed through a final linear projection layer for downstream prediction tasks.

\subsection {Data Preparation}

We evaluate the performance of our TK-GCN model in forecasting cardiac electrodynamics within a 3D ventricular geometry. The anatomical domain is discretized into 1,094 nodes and 2,184 mesh elements, forming a refined computational mesh derived from the ventricular geometry dataset provided in the 2007 PhysioNet Computing in Cardiology Challenge \citep{goldberger2000physiobank}. We simulate the propagation of cardiac electrical activity using the two-variable Aliev–Panfilov (AP) model \citep{aliev1996simple}:
\begin{eqnarray}
    \frac{\partial \omega_1}{\partial t} &=& \nabla \cdot (D \nabla \omega_1) + k_0 \omega_1 (\omega_1 - a)(1 - \omega_1) - \omega_1 \omega_2, \\
    \frac{\partial \omega_2}{\partial t} &=& \xi(\omega_1, \omega_2)\left( -\omega_2 - k \omega_1 (\omega_1 - a - 1) \right),
    \label{AP model}
\end{eqnarray}
where $\omega_1$ denotes the normalized transmembrane potential and $\omega_2$ characterizes the recovery dynamics of the electrical excitation; $D$ represents the diffusion tensor that models electrical conductivity; $ \xi(w_1, w_2) $ is a nonlinear modulation function defined as
$\xi(w_1, w_2) = e_0 + \frac{\mu_1 w_2}{w_1 + \mu_2}$;
the model parameters $a, e_0, \mu_1, \mu_2, $ and $ k_0 $ control the shape of the action potential and are adopted from \citep{aliev1996simple}. 

We employed three stimulation protocols to generate cardiac electrodynamics data under varying pathological conditions:
(1) Protocol I (Healthy Control) -- A regular pacing stimulus is applied at the apex of the ventricular geometry to simulate normal cardiac activation patterns.
(2) Protocol II (Localized Perturbation) -- An additional activation source is introduced in close proximity to the primary pacing site to induce localized interference and mimic the onset of disorganized electrical activity.
(3) Protocol III (Remote Perturbation) -- An additional activation source is placed on the right ventricle, spatially distant from the primary pacing site, to induce self-sustained and spatially discordant activation patterns, which are characteristics of fibrillatory dynamics.

We benchmark our TK-GCN model against four widely adopted methods for temporal forecasting under the three simulation protocols. (1) LSTM: An RNN architecture specifically designed to capture temporal dependencies. In our implementation, the LSTM operates in the latent space: latent representations produced by the pre-trained K-GCN encoder are input to the LSTM, and the outputs are subsequently decoded back into the original high-dimensional space via the K-GCN decoder. (2) DMD (Dynamic Mode Decomposition): A technique for forecasting future system states by approximating the dynamics in a reduced subspace. (3) Pure Koopman Operator: after encoding input data using the trained K-GCN encoder, a learned Koopman operator is applied to evolve the latent state forward in time. The resulting latent forecasts are then mapped back to the original space through the K-GCN decoder. Unlike TK-GCN, this approach does not incorporate transformer-based temporal modeling. (4) VAR (Vector AutoRegressive Model): A classical multivariate linear model that captures temporal dependencies among multiple time series. For all three stimulation protocols, we use the first 1,200 time steps as the training set and reserve the immediately subsequent 300 time steps for testing each model.

\subsection {Results of Spatiotemporal Forecasting}
\label{forcasting results}

   \begin{figure}[!ht]
	\begin{center}
		\includegraphics[width=6.5in]{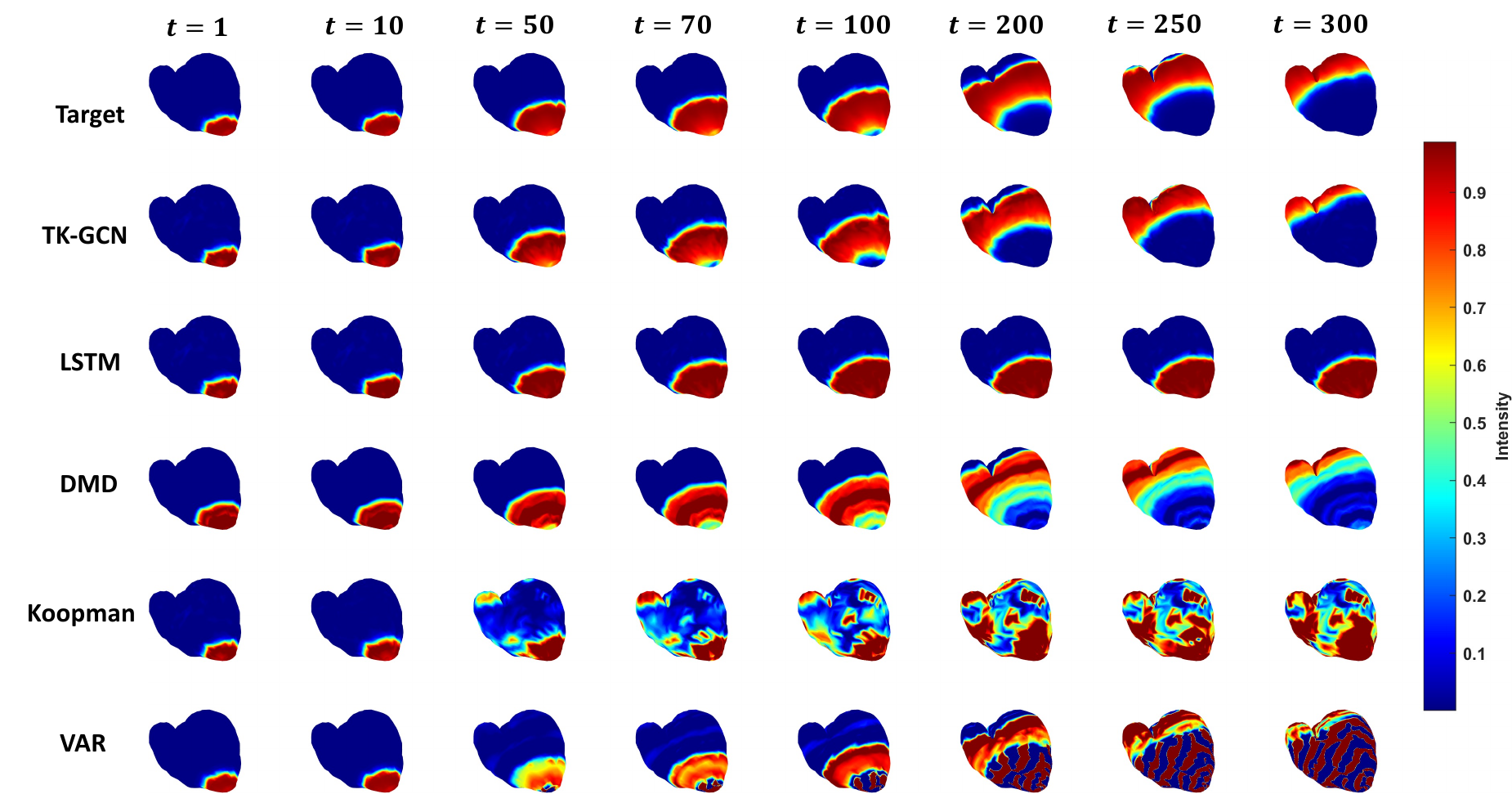}
		\caption{Predicted spatiotemporal evolution of electrodynamics under Protocol I at selected time points $(t=1, 10, 50, 70, 100, 200, 250, 300)$. The top row shows the reference targets, and the subsequent rows present predictive distributions from our TK-GCN model and benchmark methods.}
		\label{Fig:healthy HSP}
	\end{center}    
\end{figure}
We first evaluate the predictive performance of our TK-GCN model and four benchmark methods (i.e., LSTM, DMD, pure Koopman, and VAR) in capturing the spatiotemporal evolution of the dynamic distribution under three simulation protocols. The comparison is conducted at several representative time points ($t = 1, 10, 50, 70, 100, 200, 250, 300$). Fig. \ref{Fig:healthy HSP} presents the results under Protocol I, which features smooth and regular propagation patterns in the reference electrodynamics. Our TK-GCN model delivers near-perfect alignment with the reference dynamic distributions across all evaluated time points, which preserves both the sharp leading edge of the activation wave and the correct amplitude gradient from red ($\ge 0.9$) through blue ($\le 0.1$). In contrast, while both LSTM and VAR initially predict the early wavefronts accurately at $t = 1$ and $t = 10$, LSTM begins to diverge after $t = 50$: its predictions exhibit a decelerating activation front that evolves into a nearly static band by $t = 200$. VAR also captures the early propagation but progressively develops multiple anatomically implausible high-amplitude regions (“hot spots”) at later stages, disrupting the coherence of the wavefront. The DMD and pure Koopman methods initially identify the correct activation zones but deteriorate rapidly. By $t = 70$, they introduce spurious oscillations, and by $t = 200$, they yield fragmented and incoherent activation patterns. These artifacts intensify through $t = 300$, highlighting their limitations in modeling the smooth, nonlinear dynamics over longer horizons under Protocol I.

\begin{figure}[!ht]
	\begin{center}
		\includegraphics[width=6.5in]{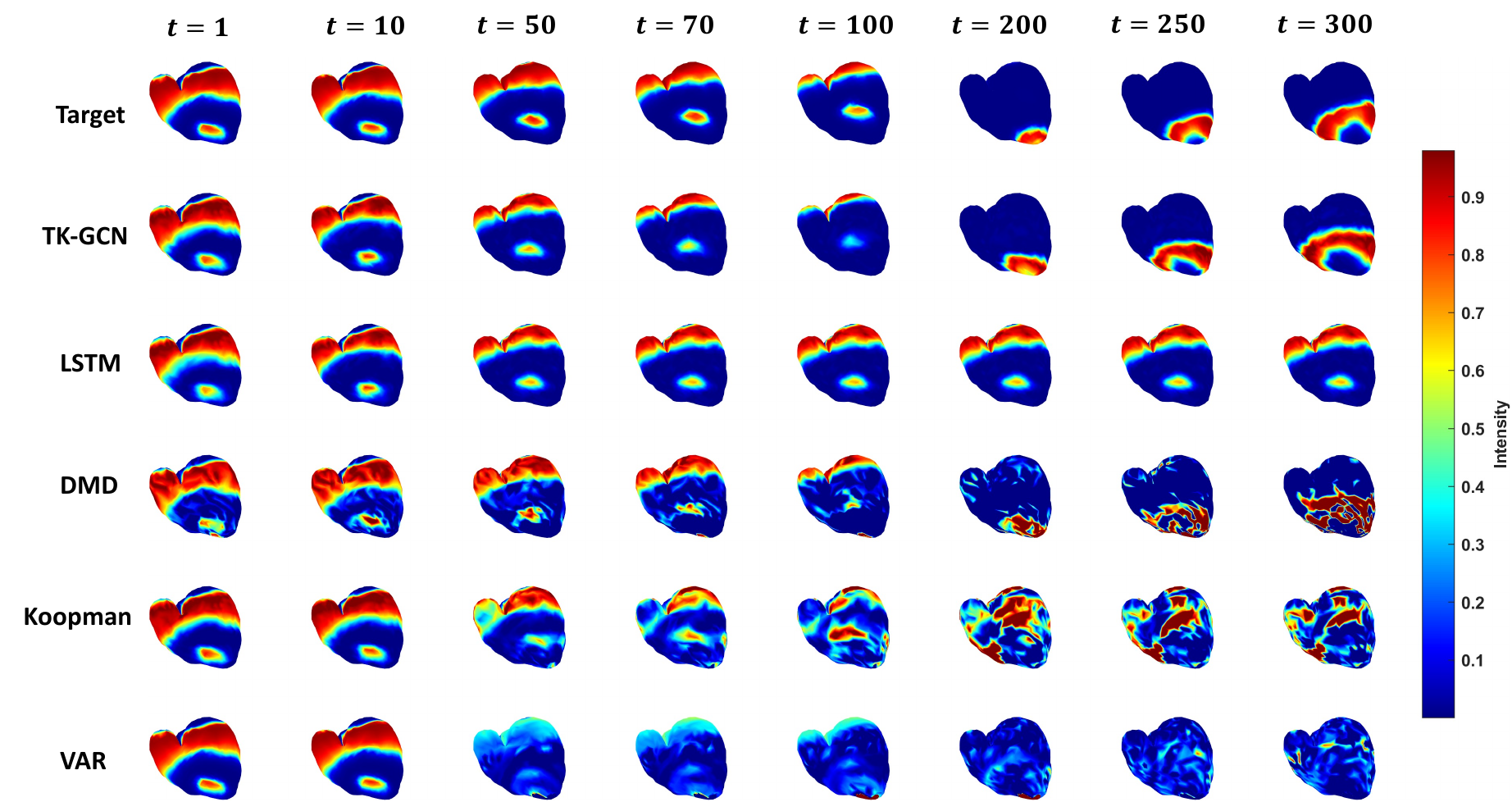}
		\caption{Predicted spatiotemporal evolution of electrodynamics under Protocol II at selected time points  $(t=1, 10, 50, 70, 100, 200, 250, 300)$. The top row shows the reference targets, and the subsequent rows present predictive distributions from TK-GCN and benchmark methods.}
		\label{Fig:unhealthy HSP}
	\end{center}    
\end{figure}

Fig. \ref{Fig:unhealthy HSP} shows the predictive performance under Protocol II with localized perturbation and largely coherent wavefronts as shown in the Target row. Similar to the results under Protocol I, our TK-GCN model consistently captures the reference activation patterns across all evaluated time points. From $t = 1$ through $t = 300$, it preserves the correct wavefront shape, amplitude gradient, and propagation speed, showing only minimal smoothing of minor irregularities at the very latest steps. In contrast, the four benchmark methods exhibit notable deficiencies. LSTM correctly identifies the early activation front at $t = 1$, $10$, $50$, and $70$, but subsequently stagnates, producing a nearly static activation band. After $t = 100$, the predicted wavefront shows little to no advancement, failing to capture the evolving spatial heterogeneity. Both DMD and pure Koopman methods approximate the early activation structures but rapidly introduce spurious high-frequency oscillations near the wavefront. These artifacts intensify over time, particularly by $t = 200$ and $t = 300$, resulting in highly fragmented activation maps that obscure the true propagation dynamics. The VAR model initially performs well at $t = 1$ and $t = 10$, but by $t = 50$ the predicted high-amplitude activation region vanishes entirely. From that point onward, the entire domain is rendered in deep blue, indicating near-zero predicted potentials across the surface. This suggests that VAR is unable to maintain or reconstruct meaningful amplitude in the presence of localized perturbations and eventually collapses to a trivial, null solution under Protocol II.

\begin{figure}[!ht]
	\begin{center}
		\includegraphics[width=6.5in]{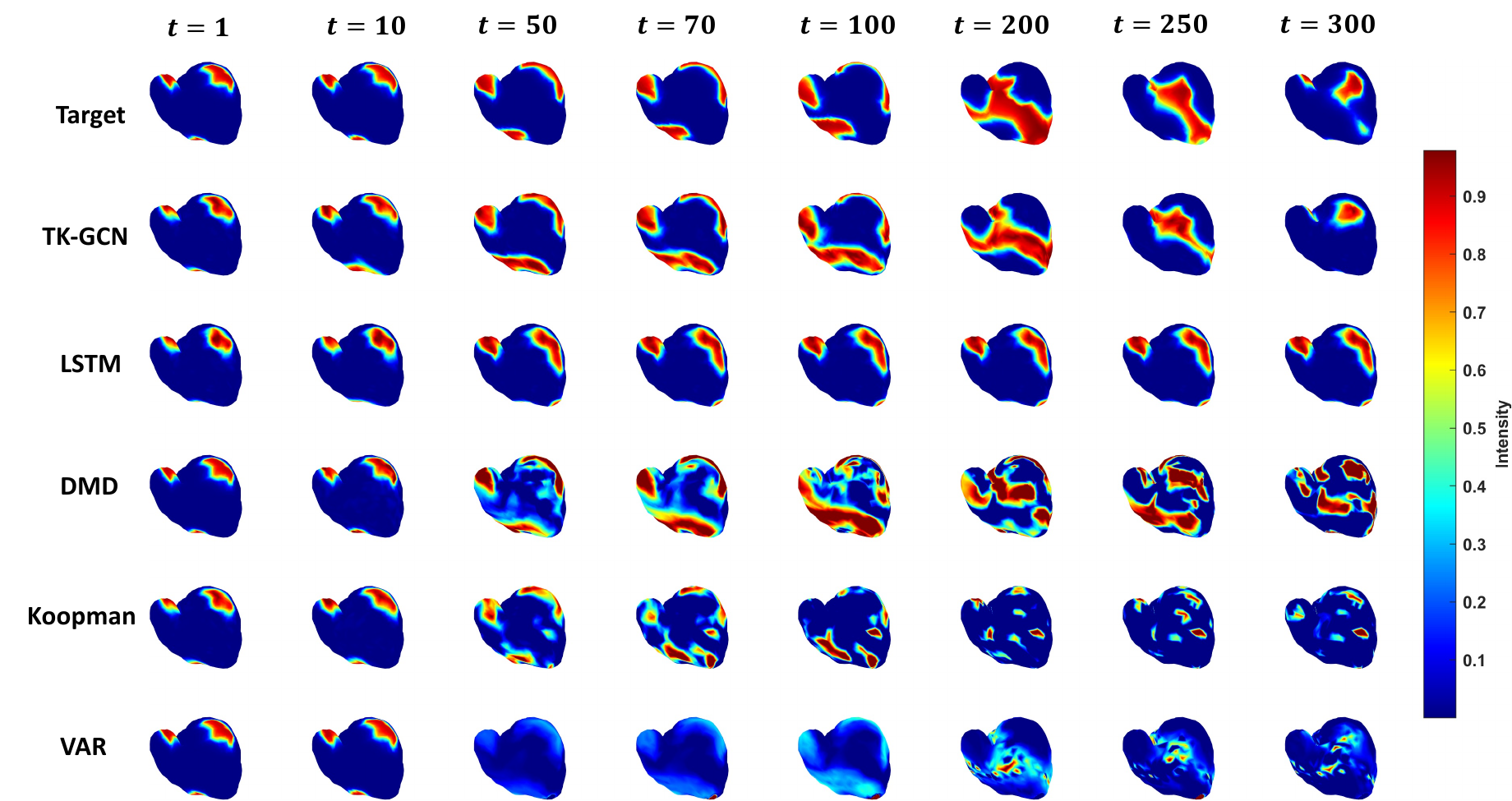}
		\caption{Predicted spatiotemporal evolution of electrodynamics under Protocol III at selected time points  $(t=1, 10, 50, 70, 100, 200, 250, 300)$. The top row shows the reference targets, and the subsequent rows present predictive distributions from TK-GCN and benchmark methods.}
		\label{Fig:unhealthy HSP 2}
	\end{center}     
\end{figure}

Fig. \ref{Fig:unhealthy HSP 2} shows the predictive performance of our third test case with pronounced wavefront fragmentation under Protocol III. Despite the increased difficulty compared to Protocols I and II, our TK-GCN model consistently outperforms all benchmark methods across the entire temporal range, demonstrating superior robustness and generalization capabilities. At early time points ($t = 1$ and $t = 10$), all models correctly localize the initial activation region; however, TK-GCN already exhibits finer spatial resolution and more accurate amplitude gradients compared to its counterparts. As the dynamics evolve ($t = 50$, $70$, and $100$), TK-GCN continues to capture the primary wavefront with high fidelity, maintaining structural coherence even as the activation propagates through highly irregular and fragmented regions. In contrast, LSTM begins to stagnate into a quasi-static band, DMD and pure Koopman degrade rapidly into incoherent oscillations, and VAR collapses into diffuse low-amplitude fields devoid of meaningful structure. At later stages ($t = 200$, $250$, and $300$), TK-GCN is the only model that faithfully reproduces the activation pattern observed in the ground truth. It accurately captures the wave trajectory, preserves the core activation region, and maintains timing consistency. Although minor smoothing at the wavefront edges is observed, the model retains the dominant activation morphology and amplitude distribution, underscoring its unique ability to model complex, long-horizon electrodynamic behavior with precision and stability.

\begin{figure}[!ht]
	\begin{center}
		\includegraphics[width=6.5in]{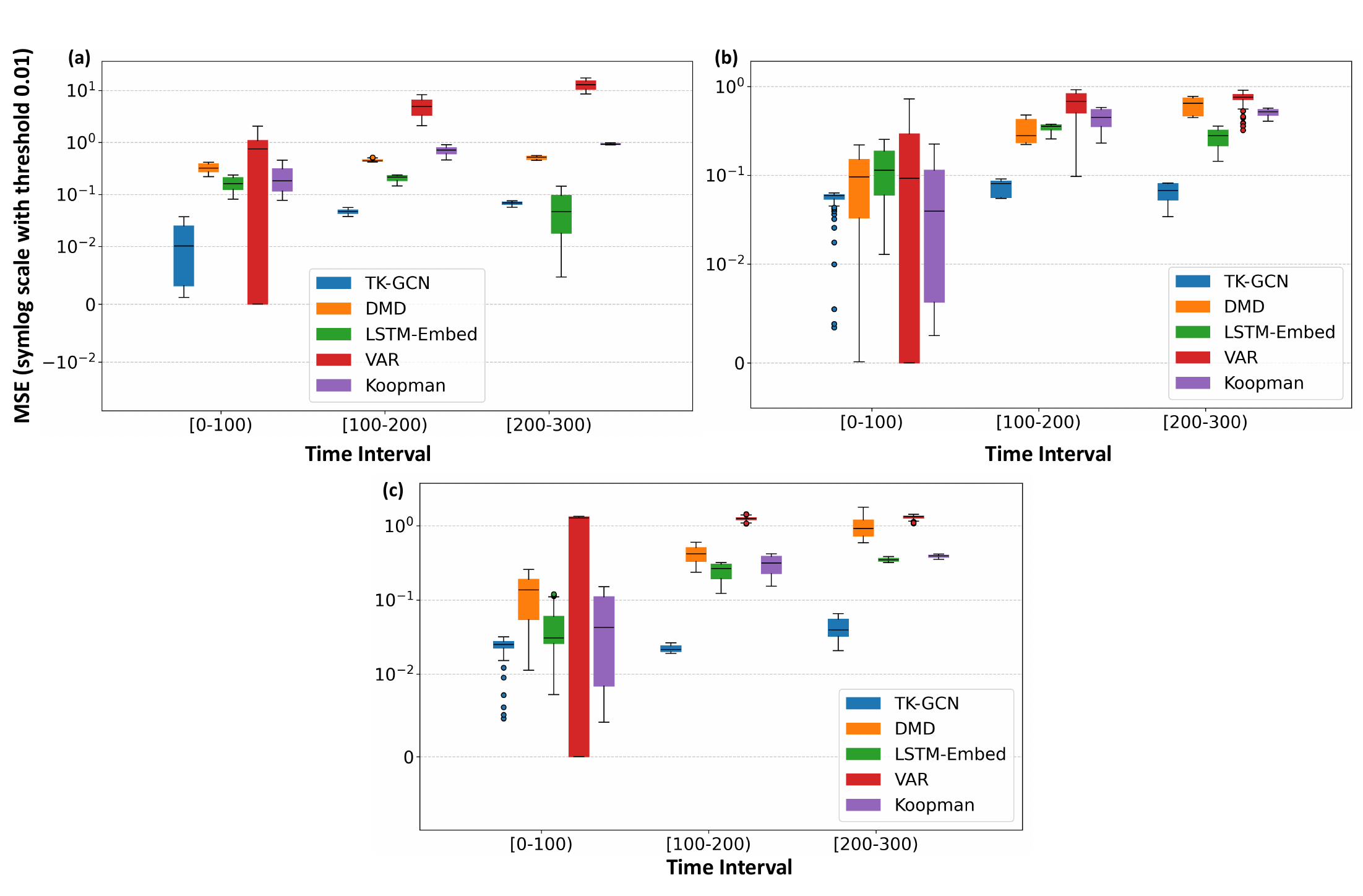}
		\caption{Box plots of MSE for TK-GCN and benchmarks at different predictive time intervals under (a) Protocol I, (b) Protocol II, and (c) Protocol III.}
		\label{Fig:boxplot}
	\end{center}    
 
\end{figure}

 Fig. \ref{Fig:boxplot} quantifies the predictive performance across all models using box plots of the mean squared error (MSE) over three forecast intervals (i.e., 0–100, 100–200, and 200–300) under each of the three simulation protocols. Under Protocol I (panel a), TK-GCN demonstrates a substantial advantage even in the earliest interval (0–100), achieving median MSE values approximately an order of magnitude lower than those of LSTM, DMD, pure Koopman, and VAR. This performance gap widens in the subsequent intervals (100–200 and 200–300), where TK-GCN consistently maintains the lowest median errors, ranging between $10^{-2}$ and $10^{-1}$, with minimal variances. In contrast, the benchmark models exhibit increasing bias and, in many cases, sharply growing variability over time, indicating reduced robustness and generalization. Similar trends are evident under Protocols II and III (panels b and c), where TK-GCN remains the top-performing model across all intervals. Even in the short-term window (0–100), the benchmark methods show significantly higher median errors and broader spreads relative to TK-GCN. These discrepancies become more pronounced in the longer forecasting horizons, highlighting the superior accuracy, temporal stability, and resilience of TK-GCN in modeling complex, heterogeneous nonlinear electrodynamics.

\subsection {Ablation Studies}
\label{ablation studies}
  The ablation studies in Table \ref{tab:ablation_tkgcn} compare the full TK-GCN model (Transformer + K-GCN), Transformer + GCN Only, Transformer Only, Transformer + CNN-based Autoencoder(AE), and LSTM + K-GCN. Performance is assessed across three forecasting intervals, i.e., [0–100), [100–200), and [200–300), under each of the three simulation protocols. Each cell reports the MSE, with boldface indicating the lowest (i.e., best) error for a given interval. By examining how errors evolve across short, mid, and long horizons, we can isolate the contributions of the Transformer backbone, the Koopman operator, and the GCN structure to overall forecasting accuracy.
  
\begin{table}[!htbp]
\centering
\caption{Ablation studies comparing TK-GCN with other model variants across three forecasting intervals}
\label{tab:ablation_tkgcn}
\renewcommand{\arraystretch}{0.55}
\begin{tabular}{llccc}
\toprule
\textbf{Dataset} & \textbf{Model} & \textbf{[0,100)} & \textbf{[100,200)} & \textbf{[200,300)} \\
\midrule
\multirow{4}{*}{Protocol I}
    & TK-GCN                    & 0.00012 & \textbf{0.00040} & \textbf{0.00062} \\
    & Transformer + GCN         & 0.00069 & 0.00171 & 0.00287 \\
    & Transformer Only          & \textbf{0.00004} & 0.00081 & 0.00143 \\
    & Transformer + CNN-AE   & 0.00005 & 0.00084 & 0.00193 \\
    & LSTM + K-GCN              & 0.00154 & 0.00179 & 0.00529 \\
\midrule
\multirow{4}{*}{Protocol II} 
    & TK-GCN                    & 0.00037 & 0.00051 & \textbf{0.00044} \\
    & Transformer + GCN         & 0.00032 & 0.00025 & 0.00051 \\
    & Transformer Only          & \textbf{0.00015} & \textbf{0.00012} & 0.00159 \\
    & Transformer + CNN-AE   & 0.00021 & 0.00031 & 0.00047 \\
    & LSTM + K-GCN              & 0.00085 & 0.00240 & 0.00182 \\
\midrule
\multirow{4}{*}{Protocol III}  
    & TK-GCN                    & \textbf{0.00017} & \textbf{0.00015} & \textbf{0.00028} \\
    & Transformer + GCN         &  0.00111 & 0.00105 & 0.00095 \\
    & Transformer Only          & 0.00252 & 0.00146 & 0.00117 \\
    & Transformer + CNN-AE   & 0.00318 & 0.00181 & 0.00157  \\
    & LSTM + K-GCN              &0.00031 & 0.00173 & 0.00238 \\
\bottomrule
\end{tabular}
\end{table}

Under Protocol I, which features smooth and periodic cardiac dynamics, the pure Transformer achieves the lowest MSE in the short-term interval [0–100), with a value of 0.00004, marginally outperforming the Transformer + CNN-AE (0.00005) and the full TK-GCN model (0.00012). This suggests that, when forecasting regular dynamic signals over short horizons, the self-attention mechanism of the Transformer alone is highly effective at capturing intrinsic temporal dependencies, while additional latent-space compression may introduce slight complexity-induced overhead without improving accuracy. However, as the prediction horizon extends to [100–200), the benefits of the full TK-GCN architecture become increasingly evident. Specifically, TK-GCN achieves an MSE of 0.00040, nearly halving the error of both the Transformer Only (0.00081) and Transformer + CNN-AE (0.00084) variants. This performance gap widens further in the [200–300) interval, where TK-GCN maintains the lowest MSE at 0.00062, while the Transformer Only and Transformer + CNN-AE models exhibit substantial error inflation, rising to 0.00143 and 0.00193, respectively. These results highlight the importance of integrating Koopman-based latent dynamics and graph-structured spatial modeling to support long-range prediction accuracy. The Transformer + GCN Only variant without latent Koopman embeddings performs significantly worse in the mid- and long-term intervals (MSEs of 0.00171 and 0.00287), underscoring the critical role of respecting the system dynamic nature for stable evolution over time. Similarly, the LSTM + K-GCN model underperforms across all intervals, with errors of 0.00154, 0.00179, and 0.00529, respectively, indicating that the limited receptive field and sequential processing of LSTM are insufficient for capturing both the local and global temporal patterns required for high-fidelity forecasting. 

The behavior under Protocol II is similar in that the plain Transformer again dominates the short term. In the [0–100) interval, Transformer Only achieves an MSE of 0.00015, closely followed by Transformer + GCN Only (0.00032) and TK-GCN (0.00037). Transformer + CNN-AE records an MSE of 0.00021, while LSTM + K-GCN is significantly higher (0.00085). As we move to the long-term interval [200–300), TK-GCN outperforms all other variants, reaching 0.00044; Transformer + CNN-AE is a close second at 0.00047, whereas Transformer Only has already ballooned to 0.00159. The fact that the pure Transformer excels up to 200 steps underscores the relatively mild chaotic behavior under Simulation Protocol II. However, once the forecasting horizon extends beyond 200, TK-GCN reduces the MSE by about 3 times compared to the pure Transformer. Overall, the results underscore that while pure Transformer architectures can handle moderately chaotic dynamics over short to mid-range intervals, the integration of Koopman operator theory and graph convolution in TK-GCN is critical for mitigating error accumulation and achieving robust long-term forecasts under the locally perturbed physiological condition.

Under Protocol III, the system dynamics are characterized by pronounced nonlinear fluctuations and fragmented wavefronts, presenting the most challenging forecasting conditions among the three protocols. In this highly chaotic setting, the full TK-GCN model clearly outperforms all other variants across all temporal intervals, achieving MSEs of 0.00017, 0.00015, and 0.00028 in the [0–100), [100–200), and [200–300) intervals, respectively. In contrast, the pure Transformer model struggles immediately with a substantially higher MSE of 0.00252 in the [0–100) interval, more than an order of magnitude worse than TK-GCN. The Transformer + GCN model offers a modest improvement over the pure Transformer, with MSEs of 0.00111 in the [0–100) interval. The limited improvement is due to the absence of Koopman-based latent dynamics, constraining its long-term accuracy. Transformer + CNN-AE performs the worst, with the highest initial error (MSE = 0.00318 in [0–100)), suggesting that CNN-based latent encoding lacks the ability to account for the complex system geometry. The LSTM + K-GCN variant performs slightly better in the short term (MSE = 0.00031), but its accuracy degrades rapidly at longer horizons (MSE = 0.00238 in [200–300)), reaffirming the limitations of sequential models in representing multi-scale, nonstationary dynamics. Collectively, these findings validate the architectural synergy in TK-GCN and its superiority in modeling multi-scale, long-horizon dynamics under diverse physiological conditions.

\section{Conclusions}

In this work, we introduce TK-GCN, a two-stage framework to tackle the fundamental challenges of spatiotemporal forecasting over irregular domains. The first stage, K-GCN, is built based on a GCN encoder–decoder architecture to learn spatially structured latent representations on arbitrary graphs. Leveraging Koopman operator theory, these latent dynamics are embedded into a near-linear manifold, facilitating more structured and interpretable temporal modeling. The second stage employs a Transformer to forecast future latent states over extended horizons, exploiting self-attention to capture dependencies that span hundreds of time steps. We evaluate TK-GCN on the task of cardiac dynamics forecasting and benchmark its performance against several baselines, including LSTM, Dynamic Mode Decomposition (DMD), a pure Koopman model, and Vector Autoregression (VAR). TK-GCN consistently outperforms all baselines across multiple forecast intervals. Ablation studies confirm that removing any of the three components, i.e., Koopman, GCN, or Transformer, causes a significant drop in prediction accuracy. These results demonstrate that combining latent linearization (via Koopman embedding), graph-based spatial smoothing (via GCN), and self-attentive temporal forecasting (via Transformer) yields stable and highly accurate long-term predictions. Additionally, the proposed TK-GCN framework can be broadly applicable to a wide range of real-world systems, including traffic networks, environmental monitoring, and biological processes, where spatial irregularity and long-range temporal dependencies are prevalent.

  \section{Data Availability Statement}
    The mesh data of the 3D geometry used in this study are openly available in the PhysioNet/Computing in Cardiology Challenge 2007 at \url{https://physionet.org/content/challenge-2007/1.0.0/}, reference number ~\citep{goldberger2000physiobank}.

\if0\blind{

\bibliographystyle{apalike}
\spacingset{1}
\bibliography{IISE-Trans}
	
\end{document}